\documentclass[runningheads]{llncs}

\usepackage{amsmath,amsfonts,bm}

\def\eqref#1{equation~\ref{#1}}

\def\1{\bm{1}}

\def\vp{{\bm{p}}}

\def\vy{{\bm{y}}}

\def\mF{{\bm{F}}}

\def\mH{{\bm{H}}}
\def\mI{{\bm{I}}}

\def\mM{{\bm{M}}}

\def\mS{{\bm{S}}}

\def\mW{{\bm{W}}}

\DeclareMathAlphabet{\mathsfit}{\encodingdefault}{\sfdefault}{m}{sl}
\SetMathAlphabet{\mathsfit}{bold}{\encodingdefault}{\sfdefault}{bx}{n}

\def\gD{{\mathcal{D}}}

\def\gF{{\mathcal{F}}}

\def\sG{{\mathbb{G}}}

\def\sR{{\mathbb{R}}}

\def\sY{{\mathbb{Y}}}

\DeclareMathOperator*{\argmin}{arg\,min}

\usepackage{eccv}

\usepackage{eccvabbrv}
\usepackage{fontawesome5}
\usepackage{multirow,makecell,graphicx}
\usepackage{wrapfig}
\usepackage{colortbl}
\usepackage{xcolor}
\usepackage{enumitem}
\usepackage{pifont}
\usepackage{subcaption}
\usepackage{hyperref}
\usepackage{url}
\usepackage{algorithm}
\usepackage{algorithmic}
\newtheorem{assumption}{Assumption}

\usepackage[capitalize]{cleveref}
\crefname{section}{Sec.}{Secs.}
\Crefname{section}{Section}{Sections}
\Crefname{table}{Table}{Tables}
\crefname{table}{Tab.}{Tabs.}
\crefname{equation}{Eq.}{Eqs.}
\crefname{algorithm}{Algorithm}{Algorithm}
\crefname{assumption}{Assumption }{Assumption }
\usepackage{graphicx}
\usepackage{booktabs}

\usepackage[accsupp]{axessibility}  

\usepackage{orcidlink}

\begin{document}

\title{DM3D: Distortion-Minimized Weight Pruning \\for Lossless 3D Object Detection}


\author{Kaixin Xu\inst{1,2}$^\ast$\orcidlink{0000-0002-7222-2628} \and
Qingtian Feng\inst{3} \and
Hao Chen\inst{1} \and
Zhe Wang\inst{1,2}$^\ast$ \and
Xue Geng\inst{1} \and
Xulei Yang\inst{1} \and
Min Wu\inst{1} \and
Xiaoli Li\inst{1} \textsuperscript{\faEnvelope[regular]}\and
Weisi Lin\inst{2} \textsuperscript{\faEnvelope[regular]}}

\authorrunning{K.~Xu et al.}

\institute{Institute for Infocomm Research (I$^2$R), Agency for Science, Technology and Research (A*STAR), 1 Fusionopolis Way, 138632, Singapore 
\email{\{xuk,chen\_hao,wang\_zhe,geng\_xue,yang\_xulei,wumin,xlli\}@i2r.a-star.edu.sg} \and
College of Computing and Data Science (CCDS), Nanyang Technological University (NTU), Singapore \\
\email{chunyun001@e.ntu.edu.sg,wslin@ntu.edu.sg} \and
National University of Singapore, Singapore \\
\email{feng\_qingtian@u.nus.edu}
}

\maketitle

\let\thefootnote\relax\footnotetext{$^\ast$ Equal contribution. \faEnvelope[regular] Corresponding author.}

\begin{abstract}
Applying deep neural networks to 3D point cloud processing has attracted increasing attention due to its advanced performance in many areas, such as AR/VR, autonomous driving, and robotics. However, as neural network models and 3D point clouds expand in size, it becomes a crucial challenge to reduce the computational and memory overhead to meet latency and energy constraints in real-world applications.
Although existing approaches have proposed to reduce both computational cost and memory footprint, most of them only address the spatial redundancy in inputs, \textit{i.e.} removing the redundancy of background points in 3D data. 
In this paper, we propose a novel post-training weight pruning scheme for 3D object detection that is (1) orthogonal to all existing point cloud sparsifying methods, which determines redundant \emph{parameters} in the pretrained model that lead to minimal distortion in both locality and confidence (detection \textbf{distortion}); 
and (2) a universal plug-and-play pruning framework that works with arbitrary 3D detection model. 
This framework aims to minimize detection distortion of network output to maximally maintain detection precision, by identifying layer-wise sparsity based on second-order Taylor approximation of the distortion. 
Albeit utilizing second-order information, we introduced a lightweight scheme to efficiently acquire Hessian information, and subsequently perform dynamic programming to solve the layer-wise sparsity.
Extensive experiments on KITTI, Nuscenes and ONCE datasets demonstrate that our approach is able to maintain and even boost the detection precision on pruned model under noticeable computation reduction (FLOPs).
Noticeably, we achieve over $\mathbf{3.89}\times, \mathbf{3.72}\times$ FLOPs reduction on CenterPoint and PVRCNN model, respectively, without mAP decrease, significantly improving the state-of-the-art.

\end{abstract}

\section{Introduction}
\label{sec:intro}

\begin{figure}[t]
    \centering
    \includegraphics[width=\linewidth]{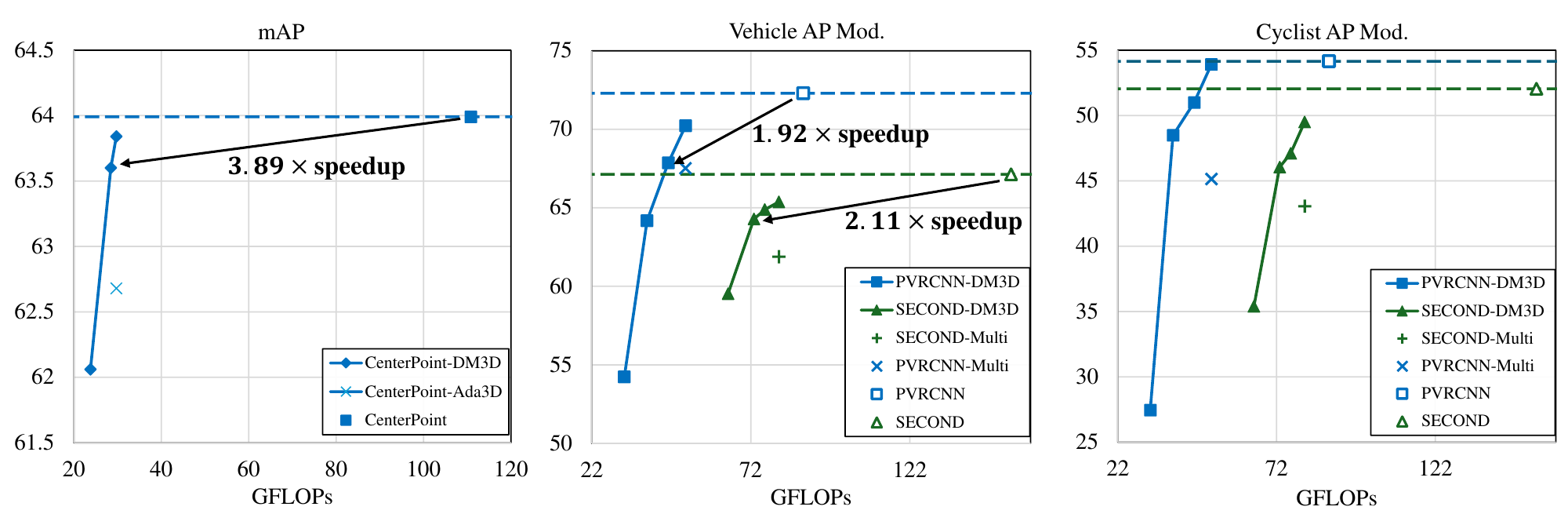}
    \caption{AP scores v.s. FLOPs on ONCE dataset of DM3D compared to baselines. }
    \label{fig:overall-once}\vspace{-10pt}
\end{figure}

3D deep learning has gained great interest from both research and industry for its wide applications in autonomous driving, robotics and VR/AR, etc. 
Particularly, 3D object detection is one of the essential visual tasks for autonomous driving systems to understand the driving environment, which serves as the foundation for the subsequent decision-making process. 
Recent advances in LiDAR-based object detection models~\cite{yan2018second,lang2019pointpillars,shi2020pv,ye2020hvnet,yi2020segvoxelnet,ao2021spinnet,yin2021center,mao2021voxel,yang20203dssd,chen2023voxelnext} show the demand of significantly costly computation to meet an empirically acceptable accuracy, particularly in handling the challenges posed by highly heterogeneous and unstructured 3D data.
However, to maximally mitigate hazardous events in real-world driving, fast inference is crucial for achieving low-latency detection.
Therefore, accelerating the object detection on 3D data with high computation and memory demand to make them more feasible in real applications becomes an urgent task. 

Past attempts~\cite{spconv2022,tang2023torchsparse++} exploited spatial sparsity in 3D point cloud modality and skipped unnecessary computation to obtain acceleration.
Other attempts further identified unimportant points and voxels from raw LiDAR data to cut down memory footprint~\cite{sun2021rsn,liu2022spatial,zhao2023ada3d}.
All of such exploitation only considers removing computation in voxel-wise connections, agnostic to explicit detection precision preservation in their formulations. 
The latest state-of-the-art sparse 3D object detection method~\cite{zhao2023ada3d} set a record of 20\% floating-point operations per second (FLOPs) reduction achieved via removing spatial redundancy.
However, there is an inevitable upper bound regarding to the FLOPs reductions when only cutting down point cloud information.
On the other hand, the redundancy in the weights of 3D models remains notably high and has not been adequately addressed.
Existing weight sparsification method for \textit{3D segmentation task}~\cite{he2022not} pruned out convolution kernel connections with least neighbouring point access rate.
However, such hit-rate-based weight selection scheme is sub-optimal on maintaining accuracy compared to typical magnitude-based and Taylor-based ranking schemes~\cite{snip,lamp,xu2023efficient} that have been proven to achieve up to over $90\%$ on benchmark datasets like CIFAR-10.
Another challenge in weight sparsification for 3D models is how to allocate layer-wise pruning ratios such that minimizes the negative impact on model accuracy. \cite{he2022not} addressed this problem by employing a greedy search on single-layer rate-distortion relations. Although they have relaxed the allocation problem into sub-problems of each consecutive two layers groups to make the solution tractable, such relaxation still requires extensive real data collections, undermining the efficiency. 

In this paper, we propose a generalized weight pruning framework for 3D object detection models.
Inspired by classic rate-distortion theory, our approach minimizes the detection distortion of 3D detection models, including the detection locality and confidence, under the constraint of computation complexity (FLOPs).
The pruning of weights is performed in a layer-wise manner where the pruning ratio of each layer is decided according to a Hessian-based rate-distortion score, which can be calculated efficiently by pre-computed gradient values.
We then develop an ultra-fast dynamic programming algorithm with polynomial time complexity to find the globally optimal pruning ratio of each layer.
Note that our proposed weight pruning and the previously proposed input pruning approaches can compensate with each other. 
By incorporating both weight and input pruning, we expect to obtain a maximal compression ratio.
We compare with previous arts and perform extensive experiments on various 3D detection models to demonstrate the effectiveness of our approach.
As shown in \cref{fig:overall-once}, we achieve remarkable speedups with neglegible performance drops or even gains on various datasets.

To the best of our knowledge, this is the first work that systematically proposes a weight pruning approach for 3D detection models in a distortion-minimized manner. We summarize the main contribution of our paper as follows:
\begin{itemize}
    \item We propose a generic weight pruning framework for 3D object detection models capable of reducing computation complexity (FLOPs) via sparsifying weights. The pruning objective is formulated as a Pareto-optimization, which explicitly minimizes distortions of both detection bounding box locality and classification confidence. The proposed pruning scheme can be applied as a standalone plug-and-play module for post-train processing for an 3D objection model, and can also serve complementary with other spatial pruning methods. 
\item Our approach adopts a hessian-based layer-wise pruning scheme. Through performing dynamic programming and fine-grained optimization, we derive an extremely efficient algorithm with polynomial time complexity to find the global optimal solution. Thanks to the distortion-minimized objective and the fine-grained optimization, our approach can maximally preserve detection accuracy under very high pruning ratio.

\item Our approach achieves state-of-the-art on six 3D object detection models at three benchmark datasets.
\end{itemize}

\section{Related Works}
Typically, neural networks for 3D point clouds primarily employ either Point-based~\cite{qi2017pointnet} or Voxel-based approaches. Point-based methods directly process the raw 3D point cloud data using neural networks, whereas Voxel-based techniques first voxelize the raw data and then operate on the voxels rather than the original data.
Numerous works~\cite{rukhovich2022fcaf3d,yan2018second,shi2020pv,deng2021voxel,yin2021center,chen2023voxelnext} took voxel-based approach for 3D object detection task.

\subsection{Spatial Sparsity}

\noindent\textbf{Sparse 3D Convolution.}
Sparse 3D Convolutional Neural Networks (CNNs) are designed to efficiently process sparse point cloud data, where a significant portion are background points with zero values. Therefore computing convolution on such data can be done by skipping empty positions in 3D space. 
\cite{engelcke2017vote3deep,riegler2017octnet} built hierarchical structures for efficient point cloud data representation.
Regular Sparse Convolution (RSC)~\cite{graham2014spatially} and Submanifold Sparse Convolution (SSC)~\cite{graham2017submanifold} were proposed to mitigate the increase of active sites in order to maintain spatial sparsity across Sparse Conv layers. 
Some libraries~\cite{tang2022torchsparse, yan2018second} have been developed to support the Sparse 3D convolution for fast inference. 
\cite{xu2020squeezesegv3} applied different filters based on the input image's spatial location, addressing the inefficiencies of using standard convolutions on LiDAR data with spatially varying distributions.

\noindent\textbf{Spatial Redundancy Reduction.}
It is easy to observe that natural redundancy in points/voxels still exist even in active regions of 3D data.
Thereby, this approach aims to further identify redundant spatial data points to reduce computational complexity. 
~\cite{graham2014spatially} was an early work proposed that output points should be omitted when corresponding input points are absent, ensuring computations are only performed for areas with relevant input data.
~\cite{guo20223d} simplified set abstraction procedure for point-based 3D action recognition models to identify important points and frames. 
\cite{Chen_2022_CVPR,liu2022spatial} proposed to engage feature sparsity by predicting importance maps to remove redundant regions, achieving computation and memory reductions. 
Multi~\cite{li2023multi} proposed a focal loss to predict voxel importance combined with voxel distillation. 
Ada3D~\cite{zhao2023ada3d} further leveraged the spatial redundancy in 2D BEV Backbone by proposing Sparsity Preserving BatchNorm to perform 3D-to-BEV feature transform. 
\cite{liu2022ins,Chen_2023_CVPR} exploited the fact that the differences (residuals) between consecutive 3D frames are typically sparse.

\subsection{Sparsity in Model Weights}
Apart from exploiting the intuitive spatial redundancy, much less attempts has been done to leverage weight redundancy in 3D networks.
~\cite{zhao2021brief} used a generator to propose pruning strategies and an evaluator that employs Bayesian optimization to select out pruning strategies for 3D object detection. They configured unique pruning schemes and rates for each layer.
Not All Neighbours\cite{he2022not} identified kernel neurons that are least frequently attended to compute output features in 3D segmentation models. They also performed a layerwise sparsity allocation by greedy search on two-layer grouping strategy.
CP$^3$~\cite{Huang_2023_CVPR} leveraged structured pruning (channel-wise pruning) for point-based 3D networks.  
None of existing weight pruning approaches is established on minimizing the impact of pruning on the detection performance as we proposed in this paper. 

\section{Preliminaries}
\noindent\textbf{Pruning within layer.}
We target at pruning learnable parameters for all feature extraction layers in 3D detection models, \textit{i.e.}, the kernels in SparseConv~\cite{spconv2022} in the 3D backbone and Conv2d in the 2D Bird's Eye View (BEV) backbone. 
To determine which parameter in a layer needs to be pruned, given a layer sparsity, we rank the neuron within each layer by the absolute first-order Taylor expansion term and eliminate the bottom ranked ones. Mathematically, we first derive the neuron ranking score matrix by the Taylor expansion $\mS = |\mW\cdot\nabla_{\mW} \vy|$~\cite{molchanov2019importance}, where $\mW$ denotes the network parameters and $\nabla_{\mW}\vy$ stands for the derivative of the network output $\vy$ to the network parameters $\mW$. 
The above pruning scheme can be formulated as $\widetilde{\mW}=\mW\odot \mM_\alpha(\mS)$, where $\mM_\alpha(\mS)$ is the binary mask generated from the ranking score matrix $\mS$ under the pruning ratio $\alpha$.
We further defines the weight perturbation $\Delta\mW = \widetilde{\mW} - \mW$ caused by a typical pruning operation to the weight. 
We further adopt a basic assumption for the weight perturbation $\Delta\mW = \widetilde{\mW} - \mW$ caused by a typical pruning operation to the weight:

\begin{assumption}\label{assum:ind-zeromean} 
    For the network with $l$ layers and the $i$-th layer  with weight $\mW^{(i)}$, \textbf{i.i.d. weight perturbation across layers}~\cite{zhou2018adaptive} which means the joint distribution across different layers is zero-meaned: 
    \vspace{-6pt}\begin{equation}
        \forall{ \  0<i\ne j<l}, E(\Delta\mW^{(i)}\Delta\mW^{(j)}) = E(\Delta \mW^{(i)}) E(\Delta \mW^{(j)}) = 0,
    \end{equation}\vspace{-6pt}
    and also zero co-variance: $E(\|\Delta\mW^{(i)}\Delta\mW^{(j)}\|^2)=0$.
\end{assumption}

\section{Methodologies}
\subsection{Problem Definition}

Given a 3D detection model consisting of (1) a feature extractor $f$ of $l$ layers with the parameter set $\mW^{(1:l)} = \big(\mW^{(1)},...,\mW^{(l)}\big)$, where $\mW^{(i)}$ is the weights in layer $i$ and (2) a detection head on top of the 2D/3D feature $f(x;\mW^{(1:l)})$ that output 3D detection bounding box predictions $\vp_b(f(x;\mW^{(1:l)}))\in\sR^{N_{s}\times S}$ and confidences $\vp_c(f(x;\mW^{(1:l)}))\in\sR^{N_{s}\times C}$ where $N_s$ denotes the number of predicted bounding boxes, $C$ stands for the number of classes and $S$ represents  the dimension of the bounding box coordinates.
Therefore the detection output is the concatenation of both bounding box prediction and class confidence scores:
\vspace{-6pt}\begin{equation}
    \vy = [\vp_b(f(x;\mW^{(1:l)}))^\top, \vp_c(f(x;\mW^{(1:l)}))^\top]^\top.
\vspace{-6pt}\end{equation}
Pruning parameters in the $f$ will give a new parameter set $\widetilde{\mW}^{(1:l)}$.
We view the impact of pruning as the distortion between the dense prediction $\vy$ and the prediction $\Tilde{\vy}$ of the pruned model. 
\vspace{-6pt}\begin{equation}
    \vy - \Tilde{\vy} = \begin{bmatrix}
     \vp_b(f(x;\mW^{(1:l)})) - \vp_b(f(x;\widetilde{\mW}^{(1:l)})) \\
     \vp_c(f(x;\mW^{(1:l)})) - \vp_c(f(x;\widetilde{\mW}^{(1:l)}))
 \end{bmatrix}.
\vspace{-6pt}\end{equation}

Given that various layers contribute to the model's performance in distinct ways~\cite{frankle2020pruning}, 
the impact of pruning layer weights would varies from layer to layer, particularly the varying information carried by active foreground points/voxels across layers. Assigning an appropriate layer-wise sparsity level for each layer could significantly impact performance.
In this regard, our proposed pruning problem is formulated as follows to obtain a layer-wise sparsity allocation that minimizes both bounding box localization distortion and confidence score distortion, constrained to a specified computation reduction target (FLOPs).
Hence we formulate a pareto-optimization problem: 
\vspace{-8pt}\begin{equation} \label{eq:obj}
 \mathrm{min.} \ E\left(\left\| \bm{\lambda}^\top (\vy - \Tilde{\vy}) \right\|^{2} \right)
 \quad s.t. \ \frac{\mathrm{FLOPs}(f(\widetilde{\mW}^{(1:l)}))}{\mathrm{FLOPs}(f(\mW^{(1:l)}))} \leq R,
\vspace{-8pt}
\end{equation}
which jointly minimizes the distortion caused by pruning under a certain FLOPs reduction target $R$. $\bm{\lambda}\in\sR_+^2$ is the Lagrangian multiplier.


\subsection{Second-order Approximation of Detection Distortion}
In order to obtain a tractable solution, the above objective needs to be transformed into some closed-form functions of the optimization variable, which is the layer-wise pruning ratio. For each layer, given a parameter scoring method, the corresponding pruning error on weight $\Delta\mW$ is also determined. 
First, we expand the distortion $\vy - \Tilde{\vy}$ using the second-order Taylor expansion (We omit the superscript $(1:l)$ for visual clarity from now)
\vspace{-8pt}\begin{equation}\label{eq:second_taylor}
    \vy - \Tilde{\vy} = \sum_{i=1}^{l}{\nabla^\top_{\mW^{(i)}} \vy \Delta\mW^{(i)} + \frac{1}{2} \Delta\mW^{(i)\top} \mH_i\Delta\mW^{(i)}},
\vspace{-5pt}
\end{equation}
where $\mH_i$ is the Hessian matrix of the $i$-th layer weight.

Then consider the expectation of the squared L2 norm in the objective~Eq.~\ref{eq:obj}, which can be rewritten as the vector inner-product form:
\begin{equation}\label{eq:expl2norm}
\vspace{-8pt}
\begin{aligned}
    &E(\|\vy - \Tilde{\vy}\|^2) = E\left[(\vy - \Tilde{\vy})^\top(\vy - \Tilde{\vy})\right] 
    = \sum_{i,j=1}^l E\left[\left(\nabla_{\mW^{(i)}}^\top \vy \Delta\mW^{(i)} \right.\right.\\ 
    &+ \left.\left.\frac{1}{2} \Delta\mW^{(i)\top}\mH_i\Delta\mW^{(i)}\right)^\top\left(\nabla_{\mW^{(j)}}^\top \vy \Delta\mW^{(j)} + \frac{1}{2} \Delta\mW^{(j)\top} \mH_j\Delta\mW^{(j)}\right)\right].
\end{aligned}
\vspace{-8pt}
\end{equation}
When we further expand the inner-product term, the cross-term for each layer pair $(i, j)$ ($1\le i\neq j \le l$) is:
\begin{equation}\label{eq:crossterm}
\vspace{-8pt}
\resizebox{\textwidth}{!}{$
\begin{aligned}
    &E\left[\Delta\mW^{(i)\top}\nabla_{\mW^{(i)}}\vy\nabla_{\mW^{(j)}}^\top \vy\Delta\mW^{(j)}\right] + E\left[\frac{1}{2}\Delta\mW^{(i)}\Delta\mW^{(i)\top}\mH_i^\top\nabla_{\mW^{(j)}}^\top \vy\Delta\mW^{(j)} \right] + \\
    &E\left[\frac{1}{2}\Delta\mW^{(i)\top}\nabla_{\mW^{(i)}}\vy \Delta\mW^{(j)\top} \mH_j\Delta\mW^{(j)}\right] + E\left[\frac{1}{4}\Delta\mW^{(i)}\Delta\mW^{(i)\top}\mH_i^\top \Delta\mW^{(j)\top} \mH_j\Delta\mW^{(j)}\right].
\end{aligned}
\vspace{-8pt}$}
\end{equation}
When we discuss the influence of the random variable $\Delta \mW$, we can treat the first-order and second-order derivatives $\nabla_{\mW}\vy$ and $\mH$ as constants and thus move them out of expectation. Also vector transpose is agnostic inside expectation. Then Eq.~\ref{eq:crossterm} becomes
\begin{equation}\label{eq:crossterm2}
\begin{split}
    &\nabla_{\mW^{(i)}}\vy\nabla_{\mW^{(j)}}^\top \vy E(\Delta\mW^{(i)\top}\Delta\mW^{(j)}) + \frac{1}{2}\mH_i^\top\nabla_{\mW^{(j)}}^\top \vy E(\Delta\mW^{(i)}\Delta\mW^{(i)\top}\Delta\mW^{(j)}) +\\
    &\frac{1}{2}\nabla_{\mW^{(i)}}\vy \mH_jE(\Delta\mW^{(i)\top}\Delta\mW^{(j)\top}\Delta\mW^{(j)}) + \frac{1}{4}\mH_i^\top\mH_j E(\|\Delta\mW^{(i)\top}\Delta\mW^{(j)}\|^2).
\end{split}
\end{equation}
Using~\cref{assum:ind-zeromean}, we can find that the above 4 cross-terms are also equal to zero, then we can derive the expectation of the distortion as follows.
\begin{equation}\label{eq:additiv}
    E(\|\vy - \Tilde{\vy}\|^2)= \sum_{i=1}^l{E\left(\left\|\nabla_{\mW^{(i)}}^\top \vy \Delta\mW^{(i)} + \frac{1}{2} \Delta\mW^{(i)\top}\mH_i\Delta\mW^{(i)}\right\|^2\right)}.
\end{equation}
After the above relaxation, we estimate the original objective as:
\vspace{-10pt}\begin{equation}\label{eq:obj_f}
\begin{split}
 \mathrm{min.} &\ \sum_{i=1}^l{E\left(\left\|\nabla_{\mW^{(i)}}^\top \vy \Delta\mW^{(i)} + \frac{1}{2} \Delta\mW^{(i)\top}\mH_i\Delta\mW^{(i)}\right\|^2\right)} \\
 s.t.&\ \frac{\mathrm{FLOPs}(f(\widetilde{\mW}^{(1:l)})}{\mathrm{FLOPs}(f(\mW^{(1:l)})} \leq R.
\end{split}\vspace{-10pt}
\end{equation}

\begin{algorithm}[t]
\renewcommand{\algorithmicrequire}{\textbf{Input:}}
\renewcommand{\algorithmicensure}{\textbf{Output:}}
\caption{Optimization via dynamic programming. }\label{algo:dp}
\begin{algorithmic}
\REQUIRE $T$: The total number of weights to be pruned. $\delta_{i,k}$: Output distortion when pruning weights in layer $i$, for $1 \leq i \leq l$ and $1 \leq k \leq T$.
\ENSURE The layerwise pruning ratios $\alpha_i^\ast$, for $1 \leq i \leq l$.
\FOR{$i$ from $1$ to $l$}
\FOR{$j$ from $0$ to $T$}
\STATE If $i=1$: $g_{1}^j \leftarrow \delta_{1,j}$, $s_{1}^j\leftarrow j$.
\STATE Else: $g_{i}^{j} \leftarrow \min \{g_{i-1}^{j-k} + \delta_{i,k}\}$, $s_{i}^{j} \leftarrow \argmin_{k} \{g_{i}^{j}\}$.
\ENDFOR
\ENDFOR
\FOR{$i$ from $l$ to $1$}
\STATE The number of weights pruned in layer $i$ is $s_{i}^{T}$. 
\STATE The pruning ratio of layer $i$ is $\alpha_i^\ast = \frac{s_{i}^{T}}{|\mW^{(i)}|}$
\STATE Update $T \leftarrow T-s_{i}^{T}$. 
\ENDFOR
\end{algorithmic}
\end{algorithm}



\subsection{Optimization Stratgy and Empirical Complexity Analysis}
Let us denote $\alpha_{i,k}$ to represent the pruning ratio at layer $i$ by $k$ weights, where $0 \leq \alpha_{i,k} \leq 1$ for all $i$ and $k$. 
In addressing the pruning problem defined by Eq. \ref{eq:obj_f}, our approach involves selecting the optimal pruning ratios to minimize the distortion as expressed in Equation \ref{eq:obj_f}. 
Suppose we have obtained a set of $\delta_{i,k}$, where $\delta_{i,k}$ represents the distortion error when pruning $k$ weights at layer $i$. 

Specifically, define $g$ as the state function, in which $g_{i}^{j}$ means the minimal distortion caused when pruning $j$ weights at the first $i$ layers.
The searching problem can be addressed by decomposing the original problem into sub-problems by the following state translation rule:
\vspace{-8pt}\begin{equation}
    g_{i}^{j} = \min \{g_{i-1}^{j-k} + \delta_{i,k}\}, \,\, where \,\,\, 1 \leq k \leq j.
\vspace{-8pt}\end{equation}
We can then achieve the optimal pruning solution by employing dynamic programming using the translation rule, as outlined in Algorithm \ref{algo:dp}, with a time complexity that is linear with respect to the model parameter size.

\begin{figure}[t]
    \centering
    \includegraphics[width=\linewidth]{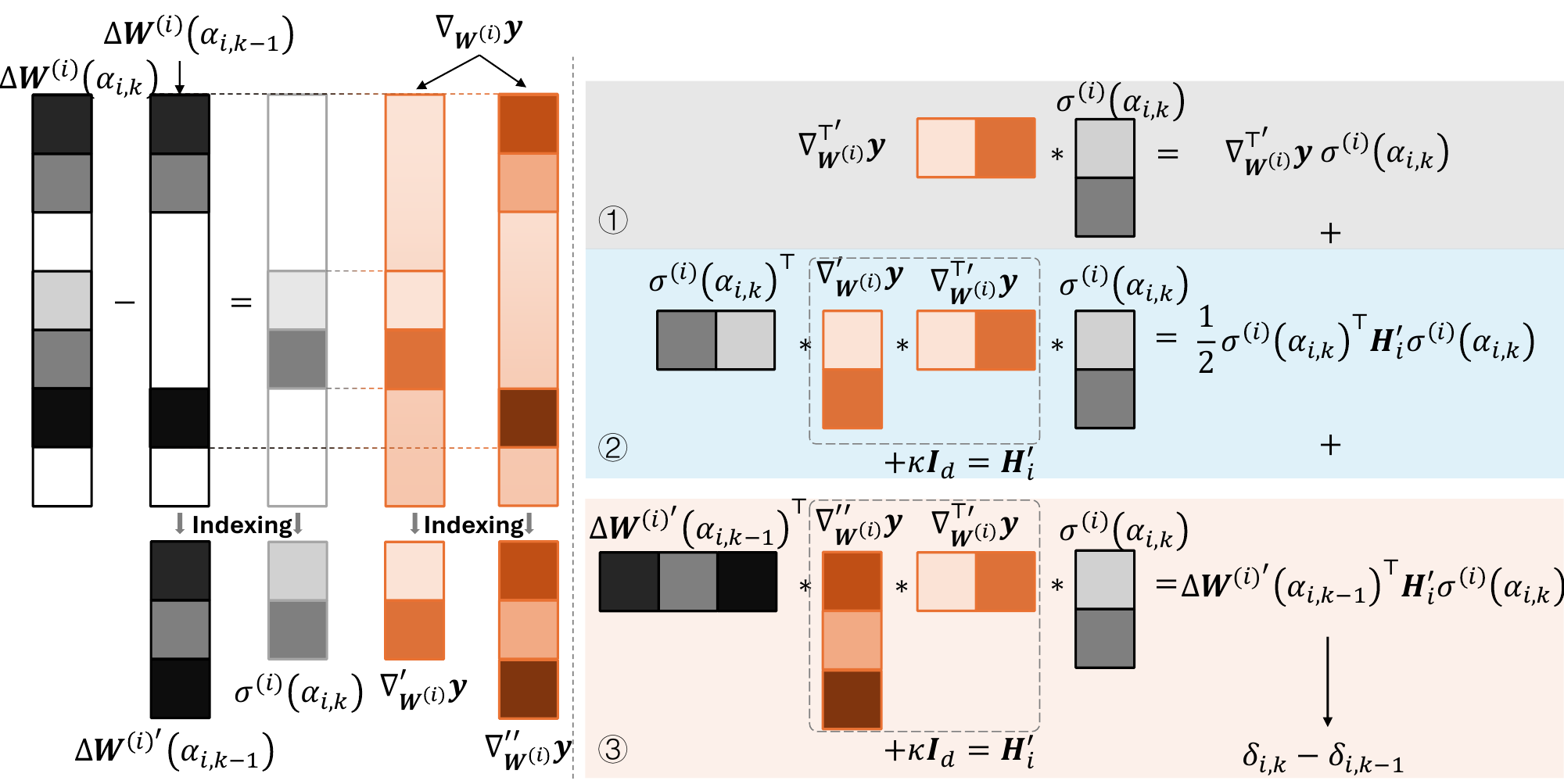}
    \caption{Detailed reduced calculation of $\delta_{i,k}$ described in Eq.~\protect\ref{eq:update}.}
    \label{fig:update}
\end{figure}

\begin{algorithm}[t]
\renewcommand{\algorithmicrequire}{\textbf{Input:}}
\renewcommand{\algorithmicensure}{\textbf{Output:}}
\caption{Distortion-Minimized Pruning of 3D Object Detection Model. }\label{algo:lda3d}
\begin{algorithmic}
\REQUIRE Training dataset $\gD_t$, Calibration dataset $\gD_c$, 3D detection model $\gF$ with $l$ layers, Number of possible pruning ratios for each layer $K$, Fine-tuning epochs $E$.
\ENSURE The pruned 3D detection model $\Tilde{\gF}$.
\STATE Inference $\gF$ on $\gD_c$ to get output detections: $\sY \leftarrow \{\vy(X) \mid \forall X \in \gD_c\}$.
\STATE Perform back-propagation on $\gF$.
\STATE Get a list of averaged gradients of each layers: $\sG = \{\nabla_{\mW^{(i)}} \vy \mid 1 \le i \le l \}$.
\FOR{$i$ from $1$ to $l$}
\STATE $\alpha_{i,0} \leftarrow 0, \delta_{i,0} \leftarrow 0$.
\STATE $\Delta\mW^{(i)} (\alpha_{i,0}) \leftarrow \bm{0}$.
\FOR{$k$ from $1$ to $K$}
\STATE $\alpha_{i,k} \leftarrow \frac{k+1}{K}$.
\STATE Prune $\mW^{(i)}$ to get $\Tilde{\mW^{(i)}}$ given $\alpha_{i,k}$ : $\widetilde{\mW}^{(i)} \leftarrow \mW^{(i)}\odot \mM_{\alpha_{i,k}}(\mS)$.
\STATE Calculate pruning error matrix: $\Delta\mW^{(i)} (\alpha_{i,k}) \leftarrow \mW^{(i)} - \widetilde{\mW}^{(i)}$.
\STATE Calculate $\delta_{i,k}$ following Eq.~\ref{eq:update}.
\ENDFOR
\ENDFOR
\STATE Obtain layerwise pruning ratios $\alpha_i^\ast$ using $\delta_{i,k}$ from \cref{algo:dp}.
\FOR{$i$ from $1$ to $l$}
\STATE Prune $\mW^{(i)}$ given $\alpha_i^\ast$ : $\mW^{(i)} \leftarrow \mW^{(i)}\odot \mM_{\alpha_i^\ast}(\mS)$.
\ENDFOR
\FOR{$e$ from $1$ to $E$}
\STATE Finetune $\Tilde{\gF}$ on $\gD_t$. 
\ENDFOR
\end{algorithmic}
\end{algorithm}\vspace{-18pt}

\subsection{Performance Optimization and Empirical Complexity}
\noindent\textbf{Hessian approximation.} For empirical networks, we approximate the hessian matrix $\mH_i$ using \emph{empirical Fisher} $\hat{\mF}$~\cite{kurtic2022optimal}:
\vspace{-10pt}
\begin{equation}
    \mH_i = \mH_{\vy}(\mW^{(i)}) \approx \hat{\mF}(\mW^{(i)}) = \kappa \mI_d + \frac{1}{N}\sum_{n=1}^N {\nabla_{\mW^{(i)}} \vy_n \nabla_{\mW^{(i)}}^\top \vy_n},
\vspace{-10pt}
\end{equation}
where $\kappa\ge 0$ is a small dampening constant, $I_d$ is the identity matrix.
A straightforward way to compute $\delta_{i,k}$ on a calibration set of size $N$ is to  iterate through different pruning ratios $\alpha_{i,k}$ to determine the corresponding $\delta_{i,k}$ values.
However, even with the use of the approximated Hessian, the process would still be computationally intensive at the complexity of $O(NKD_i^4)$, where $K$ is the number of possible pruning ratios and $D_i=|\mW^{(i)}|$ is the number of neurons in $i$-th layer weight. This poses challenge in enhancing the efficiency of the proposed method efficient to fully leverage the advantages of a sparse network.
We observe that the derivative $\nabla_{\mW_i} \vy$ remains constant with changes in the pruning ratio. This allows us to reuse the Hessian matrix for all pruning ratios, reducing the complexity to $O((N+K)D_i^2 + KD_i^4)$.
However, the existence of the biquadratic complexity makes it still excessively costly. 
Additionally, we observe that with a slight increase in the pruning ratio, only a small subset of neurons is identified for further removal from the weight tensor that has already undergone pruning. Therefore we can select a subvector ${\sigma}^{(i)}(\alpha_{i,k}) = \Delta\mW^{(i)}(\alpha_{i,k}) - \Delta\mW^{(i)}(\alpha_{i,k-1})$ each time when pruning ratio increases from $\alpha_{i,k-1}$ to $\alpha_{i,k}$ and update the $\delta_{i,k}$ from $\delta_{i,k-1}$ by the following rule:
\begin{equation}\label{eq:update}
\resizebox{\textwidth}{!}{
$\delta_{i,k} - \delta_{i,k-1} = \underbrace{\nabla_{\mW^{(i)}}^{\top\prime} \vy {\sigma}^{(i)}(\alpha_{i,k})}_{\scriptsize{\textcircled{1}}} + \underbrace{\frac{1}{2}{\sigma}^{(i)}(\alpha_{i,k})^\top\mH_i^\prime {\sigma}^{(i)}(\alpha_{i,k})}_{\scriptsize{\textcircled{2}}} 
+ \underbrace{\Delta\mW^{(i)\prime}(\alpha_{i,k-1})^\top\mH_i^\prime {\sigma}^{(i)}(\alpha_{i,k}).}_{\scriptsize{\textcircled{3}}}$
}
\end{equation}
Denote the dimension of the subvector ${\sigma}^{(i)}(\alpha_{i,k})$ as $d_{i,k} \ll D_i$ equals the number of neurons newly pruned within $\Delta\mW^{(i)}(\alpha_{i,k-1})$ compared to $\Delta\mW^{(i)}(\alpha_{i,k})$ as pruning ratio $\alpha_{i,k}$ increases from $\alpha_{i,k-1}$, the multiplication calculation in Eq.~\ref{eq:update} can be operated at lower dimensions, where $\nabla_{\mW^{(i)}}^{\top\prime} \vy \in \sR^{d_{i,k}}, \mH_i^\prime\in\sR^{D_i\times d_{i,k}}$ are subvector and submatrix indexed from the original ones. 
To further elinimate any potential confusion, we illustrate the above update rule Eq.~\ref{eq:update} in \cref{fig:update}. 
Given that $\alpha_{i,0}=0$, indicating no pruning at all, this ensures that $\delta_{i,0}=0$.
Therefore, the complexity becomes the summation of $K-1$ times of updating $O(\sum_{k=2}^{K}{d_{i,k}\sum_{k^\prime=1}^{k-1}{d_{i,k^\prime}}})$.
Since $\alpha_{i,k}$ increases linearly, the $d_{i,k}\approx \frac{D_i}{K}$, therefore, the complexity is around $O(\frac{N}{2}D_i^2)$.
Hence the total computation complexity for calculating the distortion $\delta_{i,k}$ across all $l$ layers is around $O(\frac{1}{2}\sum_{i=1}^l{D_i^2})$, significantly lower than the original complexity. 

By far, we established a weight pruning framework that is formulated to minimize distortion in 3D object detection. \cref{algo:lda3d} describes the holistic pruning procedure of the DM3D method.

\section{Experiments}

\begin{table}[t]
\centering
\caption{Performance comparison of DM3D on ONCE \textit{val} set. Gray background indicates dense model results. For baseline sparse detection results, we list the performance drop with their corresponding dense ones reported in their original papers. }\label{tab:once}
\scalebox{0.87}{
\begin{tabular}{ccc|ccc|ccc|ccc}\hline\hline
\multirow{2}{*}{Method} &  \multirow{2}{*}{\makecell{\textit{FLOPs}\\(\%)}} &  \multirow{2}{*}{\makecell{mAP\\(drop)}} &  \multicolumn{3}{c|}{Vehicle (IoU=0.7)} &  \multicolumn{3}{c|}{Pedestrian (IoU=0.5)} &  \multicolumn{3}{c}{Cyclist (IoU=0.5)} \\
                &       &         & 0-30    & 30-50   & 50-Inf  & 0-30    & 30-50   & 50-Inf  & 0-30     & 30-50   & 50-Inf  \\\midrule
\rowcolor{black!10} PointRCNN\cite{shi2019pointrcnn} & / & $28.74$ & $52.09$ & - & - & $4.28$ & - &  - & $29.84$ & - & - \\
\rowcolor{black!10} PointPillar\cite{lang2019pointpillars} & / & $44.34$ & $68.57$ & - & - & $17.63$ & - & - & $46.81$ & - & - \\
\rowcolor{black!10} SECOND\cite{yan2018second} & / & $51.89$ & $71.16$ & - & - & $26.44$ & - & - & $58.04$ & - & - \\\midrule
\rowcolor{black!10} PVRCNN\cite{shi2020pv}  & /     & $52.44$ & $87.54$ & $72.29$ & $57.22$ & $21.91$ & $20.89$ & $18.18$ & $69.8$   & $54.16$ & $36.8$  \\
Multi\cite{li2023multi} & $60.61$ & -       & $-2.85$ & $-4.42$ & $-2.81$ & $-5.89$ & $-6.76$ & $-0.58$ & $-10.73$ & $-8.12$ & $-4.77$ \\
DM3D (Ours)    & $60.61$ & $\mathbf{+2}$     & $\mathbf{+0.1}$   & $\mathbf{-2.07}$ & $\mathbf{-2.55}$ & $\mathbf{+11.61}$ & $\mathbf{+5.41}$  & $\mathbf{+0.41}$  & $\mathbf{+0.31}$   & $\mathbf{-0.25}$ & $\mathbf{-0.91}$ \\\midrule
\rowcolor{black!10} SECOND\cite{yan2018second} & /     & $51.43$   & $83.28$ & $67.13$ & $49.82$ & $26.65$ & $22.88$ & $15.58$ & $68.69$  & $52.06$ & $33.3$  \\
Multi\cite{li2023multi} & $52.54$  &  -  & $-1.49$ & $-5.35$ & $-4.03$ & $-7.84$ & $-6.24$ & $-2.98$ & $-13.32$ & $-8.83$ & $-4.69$ \\
DM3D (Ours)    & $52.54$  & $-1.6$ & $\mathbf{0.45}$  & $\mathbf{-1.75}$ & $\mathbf{0.0}$     & $\mathbf{-2.44}$ & $\mathbf{-3.91}$ & $\mathbf{-1.54}$ & $\mathbf{-1.53}$  & $\mathbf{-2.55}$ & $\mathbf{-0.18}$ \\\midrule
\rowcolor{black!10} CenterPoint\cite{yin2021center}     & -     & $64.01$ & $76.09$ &   -     &   -     & $49.37$ &   -     &   -     & $66.58$  &   -     &    -     \\
Ada3D\cite{zhao2023ada3d}      & $26.82$  & $-1.31$ & $-2.26$ &   -     &   -     & $-0.71$ &   -     &   -     & $-0.95$  &   -     &  -       \\
DM3D (Ours)    & $26.82$  & $\mathbf{-0.7}$  & $\mathbf{-0.71}$ &   -     &   -     & $\mathbf{-0.48}$ &   -     &   -     & $\mathbf{-0.94}$  &   -     &  -      \\\hline\hline
\end{tabular}}
\end{table}

\begin{table}[th]
\centering
    \caption{Performance comparison of DM3D on Nuscenes \textit{val} set.}\label{tab:nus}
    \begin{tabular}{cccc}\hline\hline
    Method         & \textit{FLOPs} (\%) & mAP (drop)  & NDS (drop)   \\\midrule
    \rowcolor{black!10} PointPillar\cite{lang2019pointpillars}     & /  & $44.63$ & $58.23$ \\
    \rowcolor{black!10} SECOND\cite{yan2018second}          & /  & $50.59$ & $62.29$ \\
    \rowcolor{black!10} CenterPoint-Pillar\cite{yin2021center} & / & $50.03$ & $60.70$ \\\midrule
    \rowcolor{black!10} \Gape[0pt][0pt]{\makecell{CenterPoint (\textit{voxel=0.1})\cite{yin2021center}}}        & /       & $55.43$ & $64.63$ \\
    \makecell{Ada3D\cite{zhao2023ada3d} (\textit{voxel=0.1})}      & $33.24$        & $54.8\ (-0.63)$ & $63.53\ (-1.1)$  \\
    \makecell{DM3D (Ours) (\textit{voxel=0.1})}       & $33.24$        & $\mathbf{55.32\ (-0.11)}$ & $\mathbf{64.36\ (-0.27)}$ \\\midrule
    \rowcolor{black!10} VoxelNeXT\cite{chen2023voxelnext}            & /       & $60.5$  & $66.6$  \\
    Ada3D\cite{zhao2023ada3d}        & $85.12$        & $59.75\ (-0.75)$ & $65.84\ (-0.76)$ \\
    DM3D (Ours)      & $85.12$        & $\mathbf{60.91\ (+0.41)}$  & $\mathbf{66.91\ (+0.31)}$ \\\hline\hline
    \end{tabular}\vspace{-8pt}
\end{table}

\begin{table}[th]
\centering
    \caption{Performance comparison of DM3D on KITTI \textit{val} set for Car class.}\label{tab:kitti}
    \begin{tabular}{c|cccc|cccc}\hline\hline
    \multirow{2}{*}{Method}       & \makecell{\textit{FLOPs}\\(\%)} & \makecell{Easy\\(drop)}   & \makecell{Mod.\\(drop)}  & \makecell{Hard\\(drop)}     & \makecell{\textit{FLOPs}\\(\%)} & \makecell{Easy\\(drop)}   & \makecell{Mod.\\(drop)}  & \makecell{Hard\\(drop)}  \\\cline{2-9}
    & \multicolumn{4}{c|}{Voxel R-CNN~\cite{deng2021voxel}} & \multicolumn{4}{c}{SECOND\cite{yan2018second}} \\\midrule
    \rowcolor{black!10} Dense   & /         & $89.44$ & $79.2$  & $78.43$        & /         & $88.08$ & $77.77$ & $75.89$ \\
    SPSS-Conv\cite{liu2022spatial}    & $73.0$    & $\mathbf{+0.28}$  & $+0.05$  & $-0.04$   & $88.31$    & $+0.21$  & $-0.11$ & $-0.15$  \\
    DM3D (Ours) & $74.36$    & $+0.04$  & $\mathbf{+0.06}$  & $\mathbf{+0.11}$  & $78.38$      & $\mathbf{+0.10}$  & $\mathbf{+0.11}$  & $\mathbf{-0.03}$  \\\hline\hline
    \end{tabular}\vspace{-8pt}
\end{table}

\subsection{Experimental Settings}
\noindent\textbf{Datasets and Models.} 
We evaluate three 3D object detection tasks to prove the effectiveness of our proposed method, namely KITTI, Nuscenes and ONCE. 
KITTI dataset includes 3712 training examples, 3769 validation examples and 7518 test examples. 
Detection targets are categorized into three classes: Car, Pedestrian, and Cyclist, with ground truth bounding boxes divided into "Easy," "Moderate," and "Hard" difficulty levels. Evaluation of detection performance employs average precision (AP) for each category, using an IoU threshold of 0.7 for cars and 0.5 for pedestrians and cyclists.
The nuScenes dataset is a comprehensive autonomous driving dataset containing 1,000 driving sequences with different modalities including LIDAR and cameras. We follow the default split, where \textit{train} split includes 700 training scenes and \textit{val} set has 150 scenes.
The ONCE dataset is a large-scale LiDAR-based collection for autonomous driving, featuring 1 million scenes with 16k fully annotated for 3D object detection, utilizing mAP for performance evaluation.
%
We perform post-train pruning followed by one round of finetuning to fully recover the performance.

\noindent\textbf{Implementation details.}
We use OpenPCDet~\cite{openpcdet2020} framework to perform finetuning. 
We use pretrained weights from official sites as possible for post-train pruning.
For others, we use our retained model in KITTI and ONCE which train 80 epochs with all default setting. We set the finetuning batch size as $64$ on 8 A100 40GB GPUs with Adam optimizer, weight decay of $0.01$ for all experiments. 
We generally set the finetune learning rates $10\times$ lower than the dense training learning rates. 
We set 30 epoch for KITTI and nuScenes, 60 epoch for ONCE. 
We search all the finetune models and find the best performance based on the metric(describe above) of each dataset.

\noindent\textbf{FLOPs Calculation.}
Since we prune both 3D and 2D backbones from the detection model, we accordingly report the FLOPs of sparsified layers in both 3D and 2D backbones. 
For baseline methods that only perform voxel sparsification in 3D backbones, we recalculate the FLOPs reduction w.r.t. 3D and 2D backbones for fair comparison.
Except for results in \cref{tab:abl-back}, as we examine the effects of pruning the detection head, we report the FLOPs w.r.t. the whole network for all three cases.

\subsection{Main Results}
We conduct extensive evaluations of our pruning method on various 3D object detection datasets including Nuscene, ONCE and KITTI. 
As we presented in \cref{tab:once}, on ONCE validation dataset, DM3D achieves \textbf{higher} detection precision for \textbf{all} three tested detectors, PVRCNN~\cite{shi2020pv}, SECOND~\cite{yan2018second} and CenterPoint~\cite{yin2021center} compared to the sparse baseline methods under the same FLOPs reduction level. 
On PVRCNN~\cite{shi2020pv} and SECOND~\cite{yan2018second} model, we outperform the voxel pruning scheme~\cite{li2023multi} on all precision metrics of class Car Pedestrian and Cyclist. We also observe a \textbf{huge performance boost} from the baseline dense model on PVRCNN on the mAP score by $2\%$.
On CenterPoint~\cite{yin2021center}, we also perform better than the current SOTA~\cite{zhao2023ada3d}.
In \cref{tab:nus}, we present the results on Nuscenes dataset. 
Again, we witness less performance drop at the same level of FLOPs reduction with the baseline w.r.t. to both mAP and NDS metrics. 
On the recent VoxelNeXT network, contrastive to the Ada3D baseline~\cite{zhao2023ada3d} who bears a $0.75$ mAP drop, we even boost the mAP on the pruned model by $0.41$. 
Comparisons on KITTI datasets are shown in \cref{tab:kitti}. 
Our method continues to achieve onpar performances with baselines on two prevailing models especially for Car AP Mod. score. On Voxel R-CNN, SPSS-Conv~\cite{liu2022spatial} performs extremely well on Car AP Easy with a $0.28$ gain from dense Voxel R-CNN, whereas our DM3D method brings slightly less but still a positive performance gain.

We demonstrate our overall detection performance on one of the datasets in \cref{fig:overall-once}. 
We compare ourselves with two baseline methods Multi~\cite{li2023multi} and Ada3D~\cite{zhao2023ada3d} on three detection networks at different FLOPs reduction levels. 
On CenterPoint, we compared the pareto-frontier of our pruning scheme to Ada3D which is the only literature that reported its mAP performance. 
We observe a $3.89\times$ FLOPs reduction from original CenterPoint model using DM3D scheme while still perform better than Ada3D. 
On SECOND and PVRCNN, we perform consistently better than baseline on different detection categories while gradually decreasing the FLOPs and achieves around $2\times$ speedup for both networks.

\subsection{Ablation Study and Discussions}

\begin{table}[t]
\caption{Ablation study when pruning only certain parts of model of SECOND on KITTI \textit{val} dataset.}\label{tab:abl-back}\vspace{-10pt}
\begin{tabular}{ccc|c|ccc|ccc|ccc}\hline\hline
\multirow{2}{*}{\makecell{\textit{3D}\\(\%)}} &
  \multirow{2}{*}{\makecell{\textit{2D}\\(\%)}} &
  \multirow{2}{*}{\makecell{\textit{Head}\\(\%)}} &
  \multirow{2}{*}{\makecell{\textit{FLOPs}\\(\%)}} &
  \multicolumn{3}{c|}{Car AP} &
  \multicolumn{3}{c|}{Ped. AP} &
  \multicolumn{3}{c}{Cyc. AP} \\
      &       &       &        & Easy & Mod.  & High &  Easy & Mod.  & High &  Easy & Mod.  & High \\\midrule
\rowcolor{black!10} /     & /     & /     & / & $88.09$ & $77.77$ & $75.91$ & $53.43$ & $48.63$ & $44.2$  & $81.8$  & $66.04$ & $62.47$ \\
47.62 & 100   & 100   & $93.2$ & $+0.14$ & $+0.17$ & $+0.06$ & $-0.42$ & $-0.77$ & $-0.39$ & $-0.2$  & $-0.33$ & $-0.03$ \\
47.62 & 67.57 & 100   & $79.22$ & $-0.41$ & $-0.33$ & $-0.52$ & $+0.5$  & $-0.22$ & $-0.12$ & $-1.65$ & $-1.61$ & $-1.22$ \\
47.62 & 67.57 & 67.37  & $64.9$ & $-0.43$ & $-0.1$  & $-0.37$ & $+0.15$ & $+0.16$ & $-0.62$ & $-1.78$ & $-1.45$ & $-1.22$ \\\hline\hline
\end{tabular}
\end{table}

\begin{table}[t]
\centering
\caption{Comparison of the proposed hessian-based pruning scheme with pruning using distortion $\delta_{i,k}$ from actual network output.}\label{tab:actual_dist}
\scalebox{0.96}{
\begin{tabular}{cc|ccc|ccc|ccc}\hline\hline
\multirow{2}{*}{Method}  & \multirow{2}{*}{\makecell{\textit{FLOPs}\\(\%)}} & \multicolumn{3}{c|}{Car AP drop}     & \multicolumn{3}{c|}{Ped. AP drop} & \multicolumn{3}{c}{Cyc. AP drop} \\
                         &                           & Easy & Mod.  & High & Easy & Mod.  & High & Easy & Mod.  & High  \\
\rowcolor{black!10}  SECOND~\cite{yan2018second} & /                         & 88.09 & 77.77 & 75.91 & 53.43  & 48.63  & 44.2   & 81.8  & 66.04 & 62.47  \\
Hessian (DM3D)     & $87.85$                     & $\mathbf{+0.14}$    & $\mathbf{+0.17}$    & $+0.06$    & $-0.42$    & $\mathbf{-0.77}$    & $-0.39$    & $\mathbf{-0.2}$    & $-0.33$   & $\mathbf{-0.03}$   \\
Actual Dist. & $87.85$                     & $-0.26$   & $+0.05$    & $\mathbf{+0.1}$     & $\mathbf{+0.51}$     & $-0.9$     & $\mathbf{-0.12}$    & $-0.78$   & $\mathbf{+0.48}$    & $-0.35$  \\\hline\hline
\end{tabular}}
\end{table}

\noindent\textbf{Controlling pruning rates of different parts in backbone.}
As shown in \cref{tab:abl-back}, we explore the possibility of pruning more and more redundant weights from different modules of the detection models. 
Start from pruning 3D backbone only, we prune out $47.62\%$ FLOPs from 3D backbone of SECOND model, resulting in a total FLOPs reduction of $93.2\%$ w.r.t. the whole network. 
Then we prune the 3D and 2D backbone together, keeping the detection head untouched. 
Finally, we also include head layers into the pruning, resulting in a total FLOPs reduction of $64.9\%$.
Intuitively, we notice that the AP scores gradually decrease as the total FLOPs decreases, but by very marginal amounts, showing that there is still a \textbf{large degree of weight redundancy} in 3D detection model.

\noindent\textbf{Second-order Distortion \emph{v.s.} True Distortion.}
To evaluate whether the proposed hessian-based distortion approximation scheme is faithful to the actual detection distortion from network output, we conducted an experiment on KITTI \textit{val} dataset using the distortion data $\delta_{i,k}$ from real network prediction $\vy$ on calibration set as originally described in \cref{eq:obj}. As shown in \cref{tab:actual_dist}, two approaches display no significant performance distinction, verifying the effectiveness of the proposed hessian approximation scheme.

\begin{figure}[t]
    \centering
    \includegraphics[width=\linewidth]{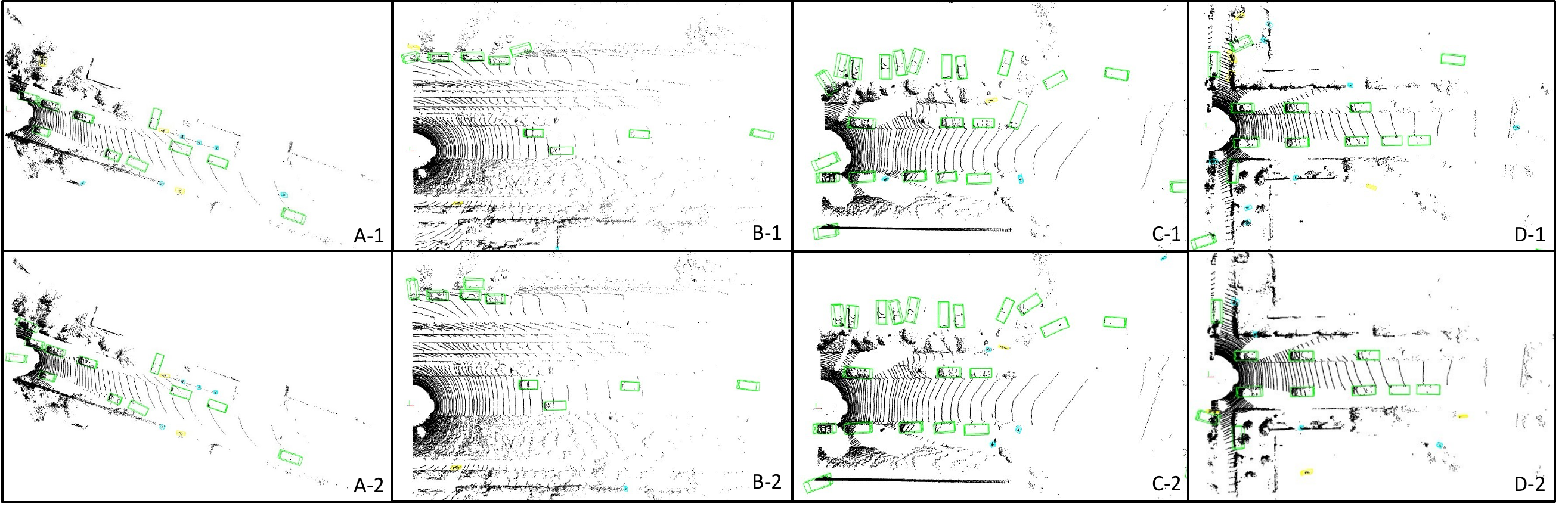}
    \caption{Qualitative performance of the pruned detection model on LiDAR data.  "A", "B", "C", "D" refer to 4 scenes from kitti dataset, of which "A" and "B" scenes are tested under the pvrcnn model while "C" and "D" scenes are tested under the second model. "1" denotes the performance of model pruned by our method and "2" denotes the performance of the pretrained model.  For example, "A-1" denotes the performance of the PVRCNN-DM3D model, "A-2" denotes the performance of the Dense PVRCNN. This figure is best viewed by zoom-in.}
    \label{fig:vis}
\end{figure}

\noindent\textbf{Qualitative Results.}
In \cref{fig:vis}, we visualize the qualitative performance of our pruning scheme applied to VoxelNeXT on Nuscenes dataset, which generates $60.91$ mAP.
One can observe that 3D detector pruned by DM3D approach generates high quality bounding boxes for essensial visual categories in LiDAR data, further verified the effectiveness of the DM3D. 

\begin{figure}[th]
    \centering\vspace{-10pt}
    \includegraphics[width=\linewidth]{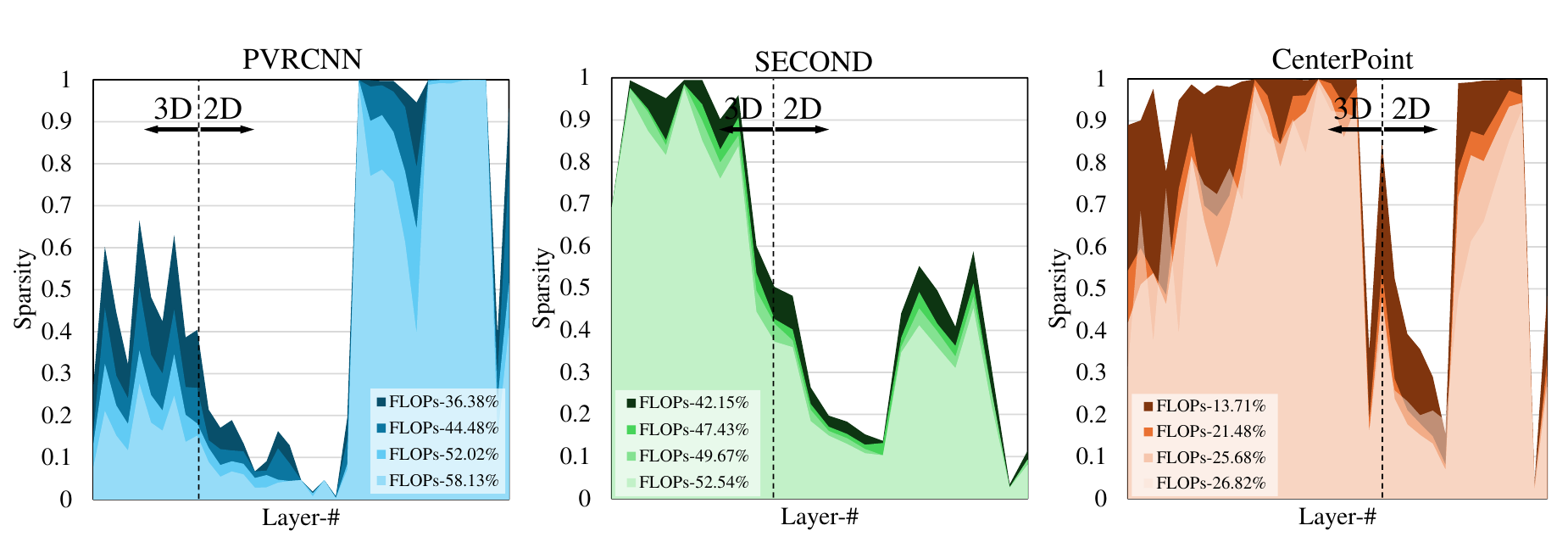}
    \caption{Layerwise sparsity allocation on three detection networks on ONCE dataset. Sparsity level close to one means most weights getting pruned out.}
    \label{fig:layerwise}
\end{figure}

\noindent\textbf{Layerwise sparsity allocation results.}
\cref{fig:layerwise} shows the detailed allocation results of the proposed DM3D. 
In \cref{fig:layerwise}, we display the layerwise sparsity levels optimized by DM3D under different FLOPs constraint levels.
We analyze the behavior on three different networks. 
Since our method leverages weight redundancies, we are able to optimize the layerwise sparsity of 3D and 2D backbones together and automatically decide the sparisty allocations.
We observe that on PVRCNN, our scheme results in more sparsity in layers in 2D backbone than in 3D part, while on SECOND, it is the opposite case where more weights in 2D backbone is preserved.
This implies that the expressiveness of PVRCNN is mostly coming from 3D feature extration than SECOND. 
Earlier layers in 2D backbone survive from most pruning cases regardless of the networks, probably because these layers are essential to smoothly transfer information from 3D domain to 2D. 
As FLOPs target decreases, the sparsity distribution remains roughly the same, where most sensitive layers regarding to the detection distortion remain in less sparsity rates.


\noindent\textbf{Weight pruning with voxel pruning.}
Our proposed weight pruning scheme works orthogonally with prevailling voxel pruning approaches. 
To explore the possibility of comining this two paradigm and potentially further reduce computations altogether, we employ the DM3D weight pruning on top of voxel pruning method SPSS-Conv~\cite{liu2022spatial} that prunes out $50\%$ of redundant voxels, and further prunes out parameters from 3D and 2D Backbones with a FLOPs reduction of $50.66\%$ \textbf{w.r.t. the whole network}. 
As shown in \cref{tab:abl-input_weight}, we observe that pruning weights from spatially sparse network brings negligible performance drop using DM3D, while the speedup is boosted from $1.36\times$ to $1.97\times$, showing the generalizability of DM3D with out-of-the-box spatial pruning methods. 

\begin{table}[t]
    \centering
    \caption{Combining spatial and weight sparsity scheme on KITTI \textit{val} dataset.}\label{tab:abl-input_weight}
    \begin{tabular}{c|cc|cccc}\hline\hline
        \multirow{2}{*}{Method} & \multicolumn{2}{c|}{Sparsity (\%)} & \multirow{2}{*}{\makecell{\textit{Total FLOPs}\\(\%)}} & \multirow{2}{*}{\makecell{Easy\\(drop)}}   & \multirow{2}{*}{\makecell{Mod.\\(drop)}}  & \multirow{2}{*}{\makecell{Hard\\(drop)}} \\
        & Voxel & Weight & & & & \\\midrule
        \rowcolor{black!10} Voxel R-CNN\cite{deng2021voxel} & $100$ & $100$ & $100$ & $89.44$ &	$79.2$ & $78.43$ \\\midrule
        SPSS-Conv\cite{liu2022spatial} & $50$ & $100$ &  $73$ & $+0.28$  & $+0.05$  & $-0.04$ \\
        SPSS-Conv\cite{liu2022spatial} + DM3D & $\mathbf{50}$ & $\mathbf{50.8}$ & $\mathbf{50.66}$ & $\mathbf{-0.04}$ & $\mathbf{+0.08}$ & $\mathbf{+0.02}$  \\\hline\hline
    \end{tabular}
\end{table}

We include more experimental details and discussions in the supplemental materials.

\section{Conclusions}
We have presented a weight pruning scheme for voxel-based 3D object detection models orthogonal to prevailing spatial redundancy-based approaches.
The established pruning scheme is based on second-order Taylor approximation on the detection distortion, which is able to minimize detection locality and confidence degradation on pruned model. 
The scheme is also extremely lightweight, with polynomial complexity for hessian information acquisition and linear complexity for layerwise sparsity searching. 
We show the superiority of the novel scheme on various 3D detection benchmarks over current state-of-the-art approaches that exploits only spatial redundancy. 
Especially, we achieved \textbf{lossless detection} with over $\mathbf{3.89}\times, \mathbf{3.72}\times$ FLOPs reduction on CenterPoint and PVRCNN model, respectively.
In the future, we aims to further formulate a unified pruning scheme that simultaneously exploits weight and spatial redundancies to enjoy the best of both worlds.

%
%
\bibliographystyle{splncs04}
\bibliography{output}

\end{document}